\documentclass[conference]{IEEEtran}
\IEEEoverridecommandlockouts
\usepackage{cite}
\usepackage{amsmath,amssymb,amsfonts}
\usepackage{algorithmic}
\usepackage{graphicx}
\usepackage{textcomp}
\usepackage{xcolor}
\usepackage{url}
\usepackage{multirow} 
\usepackage{subcaption}
\usepackage{array} 
\usepackage{bm}
\def\BibTeX{{\rm B\kern-.05em{\sc i\kern-.025em b}\kern-.08em
    T\kern-.1667em\lower.7ex\hbox{E}\kern-.125emX}}
\begin{document}

\title{ViT-LCA: A Neuromorphic Approach for Vision Transformers\\

}
\author{Sanaz M. Takaghaj}
\maketitle

\begin{abstract}
The recent success of Vision Transformers has generated significant interest in attention mechanisms and transformer architectures. Although existing methods have proposed spiking self-attention mechanisms compatible with spiking neural networks, they often face challenges in effective deployment on current neuromorphic platforms. This paper introduces a novel model that combines vision transformers with the Locally Competitive Algorithm (LCA) to facilitate efficient neuromorphic deployment. Our experiments show that ViT-LCA achieves higher accuracy on ImageNet-1K dataset while consuming significantly less energy than other spiking vision transformer counterparts. Furthermore, ViT-LCA's neuromorphic-friendly design allows for more direct mapping onto current neuromorphic architectures. 
\end{abstract}


\begin{figure*}
\centering
\includegraphics[scale=0.475]{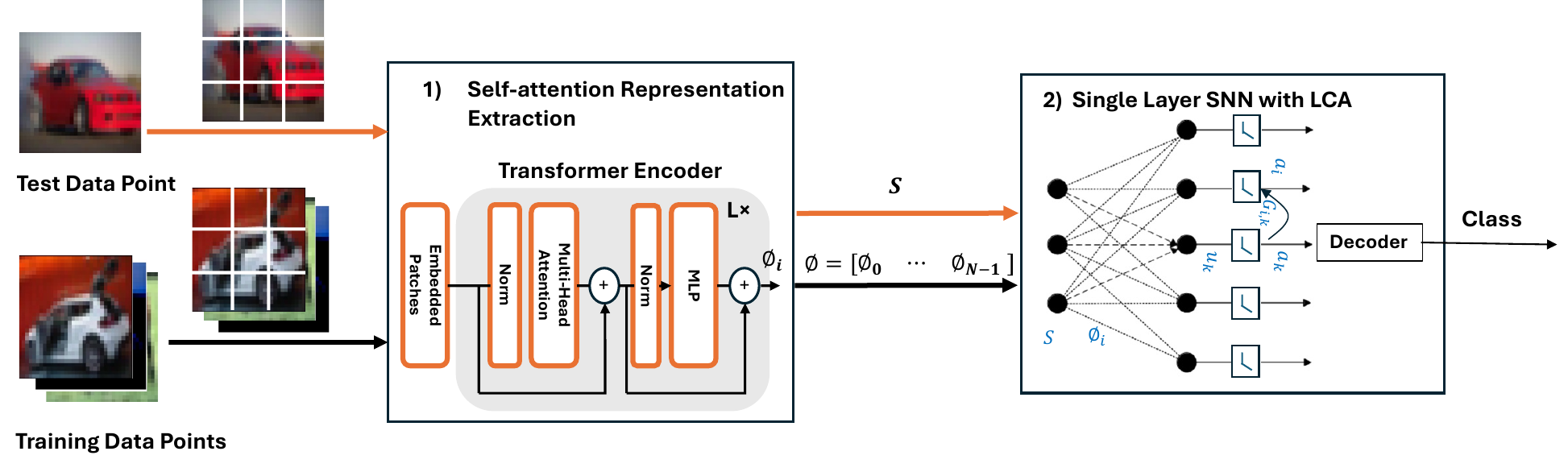}
\caption{ViT-LCA Architecture: Self-attention representations ($\phi_i$) are extracted once and stored in the synaptic weights of a single-layer SNN for inference. The orange arrows indicate the inference process that follows the completion of training.}\vspace{-12 pt}
\label{lca-arch}
\end{figure*}

\section{Introduction}
\label{sec:introduction}
Neuromorphic computing represents a paradigm shift in computing, characterized by its low-power processing capabilities and brain-inspired architectures~\cite{amir2017low,davies2018loihi,furber2014spinnaker,hoppner2021spinnaker,khaddam2022hermes,le202364,pei2019towards}. This approach emulates biological neural networks through the use of Spiking Neural Networks (SNNs). One of the primary advantages of neuromorphic chips lies in their capacity for highly parallel and energy-efficient computations. By performing operations asynchronously and maintaining proximity between synapses and weight calculations, these systems significantly reduce data movement, thereby enhancing overall computational efficiency. These platforms integrate many-core systems capable of instantiating large populations of spiking neurons, enabling information processing that mimics the dynamics of biological neural systems. Additionally, by utilizing crossbar arrays and memristores~\cite{sebastian2020memory, rao2023thousands, aguirre2024hardware} to store multi-bit quantities as conductance values, neuromorphic computing is particularly well-suited for efficiently evaluating matrix-vector-multiplications, which are fundamental to deep learning algorithms. 

A particularly interesting model in neuromorphic computing is the Locally Competitive Algorithm (LCA)~\cite{4379981, rozell2008sparse}, which is a computational model and learning algorithm that iteratively updates neuron activity to achieve a sparse representation of input data. 
This computational model has been implemented on recent neuromorphic platforms~\cite{davies2018loihi, fair2019sparse, sheridan2017sparse}. 
The competitive mechanism inherent in LCA ensures that only a limited number of neurons become active at any given time, facilitating efficient coding of high-dimensional data. One proposal for leveraging the LCA in neuromorphic computing is the Exemplar LCA-Decoder~\cite{takaghaj2024exemplar}. Functioning as a single-layer encoder-decoder, this computational model iteratively updates neuron activity to identify a sparse representation of the input data (i.e, encoding) and then uses these neuron activities for classification tasks (i.e, decoding). 

Recently, the Transformer architecture~\cite{vaswani2017attention} and its variants have demonstrated impressive performance across a range of tasks, including natural language processing~\cite{radford2018, devlin2018} and computer vision~\cite{dosovitskiy2020image,liu2021,touvron2021}. This success is largely due to their ability to effectively capture long-range dependencies, a capability primarily attributed to the self-attention mechanism.
Given the enormous computational requirements of transformer architectures, deploying these models on devices with limited resources remains a significant challenge. As a result, integrating transformer architectures with neuromorphic computing represents a promising research avenue. In particular, the combination of transformer architectures and LCA-based learning could lead to more efficient and biologically inspired artificial intelligence systems. However, this area remains largely unexplored.

This paper presents ViT-LCA, which leverages Vision Transformers (ViT)~\cite{dosovitskiy2020image} to extract self-attention representations and incorporates these representations into an LCA-based SNN. This algorithm effectively addresses the challenges of deploying transformer models on energy-constrained neuromorphic platforms. The self-attention representations are extracted once and stored in non-volatile memory elements, enabling in-memory computation on neuromorphic systems that emphasize specialized operations and energy efficiency. Our approach consists of two stages. In the first stage, a transformer encoder generates self-attention representations from the input image. In the second stage, these representations are processed by a single-layer SNN that employs a LCA encoder-decoder architecture for classification tasks. In this study, we evaluate ViT-LCA on CIFAR-10~\cite{cifar}, CIFAR-100~\cite{cifar100} and ImageNet-1~\cite{deng2009imagenet} datasets and assess the effectiveness of integrating ViT's self-attention representation with the efficiency of sparse coding through LCA for deployment on neuromorphic systems. By inputting self-attention representations (contextual embeddings) derived from ViT into a single-layer SNN model, we achieved high classification accuracy while ensuring low computational overhead and high energy efficiency.

\section{Related Work}
\label{sec:realted}
There has been a growing interest in developing methods to reduce the computational requirements of Transformer models by integrating Transformer architectures with SNNs~\cite{zhou2023, yao2023, shi2024spikingresformer, wang2023masked}, which serve as brain-inspired counterparts to deep neural networks (DNNs). Spikformer~\cite{zhou2023} and the Spike-driven Transformer~\cite{yao2023} introduce a spiking self-attention mechanism that utilizes spike-based representations for the Query, Key, and Value components, replacing traditional multiplication operations with low-energy addition operations. SpikingResformer~\cite{shi2024spikingresformer} proposes a novel Dual Spike Self-Attention mechanism to address the challenges faced by previous methods in effectively extracting local features. Wang et al.~\cite{wang2023masked} utilized the ANN-to-SNN conversion method and introduced the Random Spike Masking technique to improve the performance and energy efficiency of the Spiking Transformers.
Instead of implementing a spiking attention mechanism, ViT-LCA utilizes transformer encodings and maps them to current neuromorphic platforms.

\section{ViT-LCA}

Figure \ref{lca-arch} illustrates the architecture of ViT-LCA. The training data undergoes pre-processing before being processed by the Vision Transformer (ViT)~\cite{dosovitskiy2020image}, where it passes through attention layers to extract self-attention representations. These representations are stored in the synaptic weights of a single-layer spiking neural network using LCA encoding, which are then employed to classify unseen test inputs via a decoder.

\subsection{Extracting self-attention representations}
To extract self-attention representations from an image using the Vision Transformer (ViT), the image is first split into fixed-size patches (Tokens). Each token is then embedded into a vector representation, which is augmented with positional encodings to retain spatial information. An additional learnable ``classification token'' is appended to the sequence of tokens. The resulting sequence of embeddings is then input into the Transformer encoder. The extracted self-attention representations, denoted as $\phi_i$, are subsequently used to construct a dictionary $\phi$ as in Eq.~\ref{dictionary}.

\subsection{Single-Layer SNN with LCA Encoder and Decoder}
\label{sec:design}
We first provide an overview of the Exemplar LCA-Decoder algorithm~\cite{takaghaj2024exemplar}, which utilizes the sparse coding algorithm~\cite{rozell2008sparse} and the LCA algorithm~\cite{4379981} to represent the input signal $\mathit{S}$:
\begin{equation}
\label{recon}
S = \sum_{i=0}^{M-1} \phi_{i} a_{i}  + \varepsilon
\end{equation}
where $\phi$ is a dictionary of self-attention representations ($\phi_i$) and $a_{i}$ represents the activation of the LIF neuron $i$. The term $\varepsilon$ represents Gaussian noise. The membrane potential $u_i$ of the LIF neuron is governed by a driving excitatory input $b_i$ and an inhibition matrix (Gramian) $G$. The Gramian matrix enables stronger neurons to inhibit weaker neurons from activating, leading to a sparse representation.
\begin{equation}
\label{neuron update}
\tau \dot{u}_i[k]+u_i[k]=b_i-\sum_{m\neq i}^{M-1}G_{i,m}a_m[k]
\end{equation}

\begin{equation}
\label{b_i}
b_i = S\phi_i
\end{equation}

 \vspace{-6pt}
\begin{equation}
\label{grammian}
G = \phi^{T}\phi
\end{equation}

 \vspace{-6pt}
\begin{equation}
\label{dictionary}
\phi = [\phi_0, \phi_1, \ldots, \phi_{M-1}]
\end{equation}

Each $\phi_i$ represents a self-attention representation learned from a data point in the training dataset. This approach eliminates the need for dictionary learning, differing from the original LCA, where Stochastic Gradient Descent (SGD) is used to learn and update the dictionary for each batch of input data. The thresholding function is:
\begin{align}
\label{thresholding}
    a_i[k] &= T_\lambda(u_i[k]) \nonumber \\
    &=
    \begin{cases}
        u_i[k] - \lambda \operatorname{sign}(u_i[k]), & \quad \left| u_i[k] \right| \geq \lambda \\
        0, & \quad \left| u_i[k] \right| < \lambda
    \end{cases}
\end{align}

Here, the threshold $\lambda$ refers to the level that the membrane potential must exceed for the neuron to become active. Next, a decoder is designed to decode the sparse codes $a_i$ and map them to \textit{C} distinct classes from the training dataset. Later, the same decoder is used to generate class predictions for the unseen test data points. 

Next, any (unseen) test data $S = S_{test}$ is mapped to the resulting \textit{M}-dimensional space of self-attention embeddings (dictionary atoms $\phi_i$). The corresponding activation codes $a_i$ that approximate the new input are then determined, as outlined in Eq.~\ref{neuron update} to Eq.~\ref{thresholding}. 

Given a sufficient number of training data points \textit{M}, each test data point will have a sparse coding representation $a = [a_0, a_1, a_2, ..., a_{M-1}]$, with the majority of the $a_{i}$ values being zero. Ideally, for any test input $S_{test}$, the non-zero entries in the vector $a$ would correspond exclusively to the dictionary atoms $\phi_i$ associated with class \textit{c}, where \textit{c} ranges from 1 to \textit{C}. However, modeling errors and input signal noise may introduce small non-zero entries that are associated with multiple classes. To address this issue and better harness the relevance of neuron activations for each class, the ``Maximum Sum of Activations'' decoder is proposed. In this approach, the $\ell_1$ norms of the activations for each class \textit{c} are summed and the class with the highest value is determined:

\begin{equation}
\label{max-sum-decoder}
Predicted \; Class = \underset{c}{argmax}(\sum_i \left |  a_{i}^{(c)}\right |)
\end{equation}

\begin{table*}
  \centering
  \caption{Top-1 Test Accuracy Scores and Workload/ Energy Estimates for ViT-LCA.}
  \vspace{-8pt}
    \begin{tabular}{|c|c|c|c|c|c|c|}
    \cline{6-7}
        \multicolumn{4}{c}{} & \multicolumn{1}{c}{} & \multicolumn{2}{|c|}{\textbf{Accuracy}}  \\
    \cline{6-7}
    \multicolumn{4}{c}{} & \multicolumn{1}{c}{} & \multicolumn{2}{|c|}{Decoding Methods}  \\
      \hline
      \textbf {Dataset}& \textbf{Transformer Model} & \textbf{Training TFLOPs$^{\mathrm{a}}$} & \textbf{Inference GFLOPs$^{\mathrm{b}}$} & \textbf{Energy} & \textbf {$\underset{i}{max} \left\{ a_{i} \right\}$} & \textbf {$\underset{c}{max} \sum_{i} \left |  a_{i}^{(c)}\right |$} \\
    \hline
       CIFAR-10  & ViT-B/16 & 1.92 & 2.08 & 0.19 mJ & 90.53\%
 & \textbf{95.63\%} \\
      \hline   
    CIFAR-100& ViT-B/16 & 1.92 & 2.08& 0.19 mJ & 81.63\%
 & \textbf{81.80\%} \\
      \hline
     ImageNet-1K& ViT-B/16 & 1.92 & 2.08 & 0.19 mJ & 71.51\% & \textbf{80.24\%} \\
     \hline
\multicolumn{7}{l}{$^{\mathrm{a}}$Tera FLOPs, $^{\mathrm{b}}$Giga FLOPs}\\
\end{tabular}\vspace{-8pt}
\label{tab:decoders}
\end{table*}


\begin{table}
\caption{Hyperparameters}\vspace{-8pt}
\label{parameter}
\begin{center}\small
\begin{tabular}{|c|c|c|}
\hline
\textbf{Symbols} & \textbf{Description} & \textbf{Value} \\
\hline
- & Patch size & 16*16 \\
$L$ & Self-attention layers & 12 \\
$M$ & Dictionary size & 50K \\
$\hat{M}$ & Average number of spiking neurons& 200\\
$N$ & Self-attention
representation length & 768 \\
$\lambda$ & Threshold & 2 \\
$\tau$ & Leakage term &  100\\
$K$ & Time steps &  100\\
\hline
\end{tabular}\vspace{-8pt}
\end{center}
\end{table}

\section{Experiment Setup}
\label{sec:evaluation}
We evaluated ViT-LCA using PyTorch~\cite{pytorch} and the following datasets: CIFAR-10~\cite{cifar}, CIFAR-100~\cite{cifar100} and ImageNet-1K~\cite{deng2009imagenet}. All datasets and the pre-trained Vision Transformer model were obtained using Torchvision~\cite{torchvision}. 


\subsection{Dataset Pre-Processing}
After resizing the images to a uniform size of 224 x 224 pixels, the dataset is normalized to have a zero mean and a standard deviation of one to prepare it for input into the Vision Transformer. 

\subsection{CIFAR-10 and CIFAR-100}
The CIFAR-10 and CIFAR-100 datasets each comprise a total of 60,000 images, with 50,000 designated for training and 10,000 for testing. Each image is represented in RGB format and has dimensions of 32 x 32 pixels. The CIFAR-10 dataset contains 10 classes, while the CIFAR-100 dataset is more complex, featuring 100 classes.

\subsection{ImageNet-1K}
The ImageNet-1K dataset comprises approximately 1.28 million training images and 50,000 validation images. For the purpose of constructing the dictionary, the training dataset was randomly split, and a subset of 50,000 samples was selected. The ILSVRC2012 validation dataset was exclusively used for testing, ensuring that it was not utilized in any capacity prior to evaluation. This dataset contains 1000 classes.

\section{Evaluation}

Table~\ref{tab:decoders} presents the workload and energy estimates, along with accuracy performance across all tested datasets. All experiments were conducted using the hyperparameters specified in Table~\ref{parameter}, which were selected to achieve nearly 100\% training accuracy on all datasets; the results reflect the corresponding test accuracy. Further optimization of these hyperparameters per dataset may enhance accuracy even further. Additionally, Table~\ref{tab:decoders} indicates that the "Maximum Sum of Activations" decoder outperforms the "Maximum Activation" decoder across all datasets.  
The energy estimates were calculated based on the floating-point operations performed during inference and the anticipated energy consumption per floating-point operation, which is discussed further in the following section.

\subsection{Workload and Energy Efficiency Analysis}
\label{Workload}

In this section, we conduct a comprehensive evaluation of the workload and energy efficiency of ViT-LCA. We assess workload efficiency by estimating the number of Floating-Point Operations (FLOPs) involved, followed by a detailed discussion of energy consumption.

Computing $b_i$ as described in Eq.~\ref{b_i} requires $N*M$ multiplications and $(N-1)*M$ addition, where $N$ is the size of the self-attention representation and $M$ is the size of the dictionary (which also corresponds to the number of neurons). The inhibition signal calculation in Eq.~\ref{neuron update} involves $M^2-M$ multiplications and $M^2 -2M$ additions. Additionally, the leakage term introduces $M$ multiplication operations, while $3M$ additions are required to combine these terms and update the neurons' membrane potentials, as specified in Eq.~\ref{neuron update}. Furthermore, computing the Gramian matrix $G$ in Eq.\ref{grammian} entails $\frac{M(M+1)N}{2}$ multiplications and $\frac{M(M+1)(N-1)}{2}$ additions.

\begin{align}
\label{FLOPS-training}
\underset{(Training)}{FLOPs} = \frac{M(M+1)(2N-1)}{2}
\end{align}

The Gramian matrix computation, which is considered part of the training cost and performed once per task (dataset), is excluded from the inference cost. Thus, the total floating-point operations required per time step $K$ for inference is given by:
\begin{align}
\label{FLOPS1}
\underset{(Inferenec)}{FLOPs} = K(\frac{(2N-1)M}{K}+2M^2+M)
\end{align}
Note that $b_i$ is computed once per input data and remains constant across iterations. Finally, the expected sparsity through LCA is factored in by redefining $M$, the number of active neurons, as $\hat{M}$, which represents the average number of spiking neurons whose $a_m \neq 0$.

\begin{align}
\label{FLOPS2}
\underset{(Inferenec)}{FLOPs} = K(\frac{(2N-1)M}{K}+2M\hat{M}+M)
\end{align}

The estimated training and inference FLOPs for ViT-LCA are presented in Table~\ref{tab:decoders}. ``TFLOPs" denotes the Tera FLOPs required for training (computing the Gramian matrix), while ``GFLOPs" represents the Giga FLOPs for inference operations. Training FLOPs accounts for all training data points, whereas inference FLOPs is estimated for a single test input. 

In addition to enhancing workload efficiency, ViT-LCA leverages recent advancements in in-memory computing through memristive crossbar arrays. 
Memristive crossbars can perform multiplication operations based on Kirchhoff's Current Law and Ohm’s Law (I = V · G, where I is the current, V is the input voltage, and G is the conductance of each memristor), thereby enhancing energy efficiency and minimizing the area footprint. Using Resistive Random Access Memory (RRAM) crossbar arrays, Yao et al.~\cite{yao2020fully} reported an energy efficiency of 11 Tera OPs per Watt for floating point multiply-and-accumulate (MAC) operations. This is equivalent to each floating point operation consuming at most approximately $9.09 \times 10^{-14}$ joules of energy. 

\begin{figure}
\centering
\includegraphics[width=0.45\textwidth]{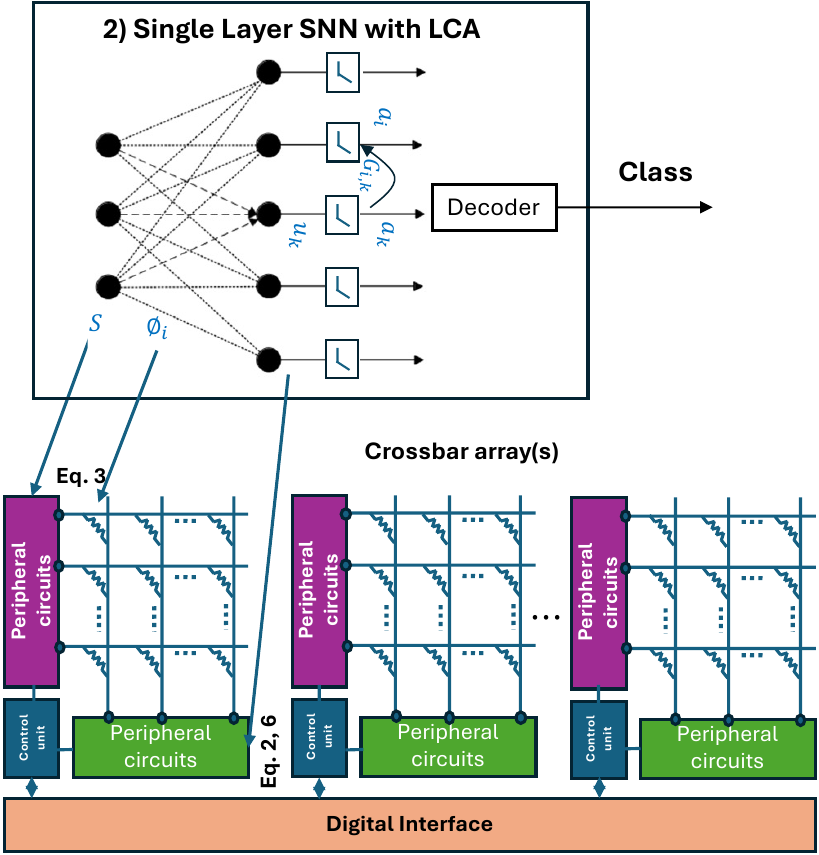}
\caption{ViT-LCA Hardware Deployment}\vspace{-8pt}
\label{fig:p}
\end{figure}

Fig.~\ref{fig:p} illustrates the hardware mapping of ViT-LCA, where the \(\phi_{i}\) values are represented as the conductance of memristors in each column, used to perform neuron excitatory input multiplications as shown in Eq.~\ref{b_i}.
Inputs are preprocessed in the input peripheral circuits, while neuron dynamics and thresholding—defined in Eq.\ref{neuron update} and Eq.\ref{thresholding}—are implemented in the output peripheral circuits located in the neural cores. The calculation of the Gramian matrix, which serves as the training head in this algorithm, can be performed offline, with the results stored in a lookup table (LUT) in the on-chip memory of a host processor that interacts with the neural cores via a digital interface.\vspace{-6pt}

\section{Discussion}
\label{}
This study serves as a proof of concept, demonstrating the potential for a uniform algorithm applicable across various transformer architectures and datasets. However, it currently exhibits limitations in accuracy compared to state-of-the-art transformer models, and further improvements could be realized by exploring alternative transformer architectures for self-attention representation extraction. 
We utilized the ViT-B/16 transformer model developed by Dosovitskiy et al.\cite{dosovitskiy2020image}, which has reported accuracies of 98.13\%, 87.13\%, and 77.91\% on CIFAR-10, CIFAR-100, and ImageNet-1K, respectively. In comparison, our results (Table\ref{tab:decoders}) were slightly lower on CIFAR-10 and CIFAR-100, but we achieved a higher accuracy on ImageNet. We believe this discrepancy arises from their use of a higher resolution of 384 and fine-tuning on CIFAR-10 and CIFAR-100, whereas our model was pre-trained solely on ImageNet-1K without fine-tuning on the CIFAR-10 and CIFAR-100 datasets, which we opted for to avoid additional computational overhead.

In comparisons with spiking transformers~\cite{zhou2023,wang2023masked,shi2024spikingresformer}, our model demonstrated superior performance across all three datasets, achieving higher accuracy than Spikformer while consuming only 0.19 mJ of energy. Spikformer recorded accuracies of 95.51\% on CIFAR-10, 78.21\% on CIFAR-100, and 74.81\% on ImageNet-1K, with an estimated inference energy of 21.48 mJ~\cite{shi2024spikingresformer}. Additionally, when compared to the Masked Spiking Transformer~\cite{wang2023masked} based on Swin transformer~\cite{liu2021swin}, our model achieved higher accuracy on ImageNet, but showed lower accuracy on CIFAR-10 and CIFAR-100, where the Masked Spiking Transformer recorded accuracies of 97.06\%, 86.73\%, and 77.88\%, respectively, with time steps of 128. SpikingResformer~\cite{shi2024spikingresformer} achieved Top-1 accuracy of 79.40\% on ImageNet-1K, with an energy consumption of 14.76 mJ. With the highest accuracy results and lowest energy consumption on ImageNet, along with direct deployment on neuromorphic systems, ViT-LCA demonstrates significant potential for future implementation on neuromorphic platforms.
\vspace{-2pt}

\section{Conclusion}
\label{sec:conclusion}
\vspace{-2pt}
In this work, we explored the integration of LCA with Vision Transformers (ViT) for the first time and evaluated its performance on classification tasks using the CIFAR-10, CIFAR-100, and ImageNet-1K datasets. We leveraged self-attention representations extracted through ViT within an LCA encoder-decoder framework, eliminating the need to train a dictionary as required in the original LCA implementation. Additionally, we demonstrated how this approach can be effectively mapped to memristor crossbar arrays to leverage efficient in-memory computing on neuromorphic systems. This study underscores the potential for further research, including the investigation of additional transformer architectures to achieve even higher accuracy results.


\vspace{12pt}

\end{document}